\documentclass{article}

\usepackage[preprint,nonatbib]{neurips_2021}
\usepackage[numbers,sort&compress]{natbib}

\usepackage[utf8]{inputenc} 
\usepackage[T1]{fontenc}    
\usepackage{hyperref}       
\usepackage{url}            
\usepackage{booktabs}       
\usepackage{amsfonts}       
\usepackage{nicefrac}       
\usepackage{microtype}      
\usepackage{xcolor}         

\usepackage{subfigure}
\usepackage{setspace}
\usepackage{multirow}
\usepackage{graphicx}
\usepackage{tikz}
\usepackage{bbm}
\usepackage{mathtools}
\usepackage{makecell}
\usepackage{colortbl}
\usepackage{amsmath,amssymb,bm}

\usepackage{multicol}
\usepackage{wrapfig}

\title{One Model, Multiple Modalities:  A Sparsely Activated Approach for Text, Sound, Image, Video and Code}
\author{
  Yong Dai$^*$, Duyu Tang\thanks{Correspondence to: Duyu Tang (\texttt{duyutang@tencent.com}), $^*$ indicates equal contribution}\ \ ,  Liangxin Liu$^*$, Minghuan Tan$^*$,  Cong Zhou$^*$, Jingquan Wang$^*$, \\
  \textbf{Zhangyin Feng, Fan Zhang, Xueyu Hu, Shuming Shi} \\  \\ \\
  Tencent AI Lab
  }

\begin{document}
\maketitle
\begin{abstract}
People perceive the world with multiple senses (e.g., through hearing sounds, reading words and seeing objects). However, most existing AI systems only process an individual modality. This paper presents an approach that excels at handling multiple modalities of information with a single model. In our ``\textbf{SkillNet}'' model, different parts of the parameters are specialized for processing different modalities. Unlike traditional dense models that always activate all the model parameters, our  model sparsely activates parts of the parameters whose skills are relevant to the task. Such model design enables SkillNet to learn skills in a more interpretable way. We develop our model for five modalities including text, image, sound, video and code. Results show that, SkillNet performs comparably to five modality-specific fine-tuned models. Moreover, our model supports self-supervised pretraining with the same sparsely activated way, resulting in better initialized parameters for different modalities. We find that pretraining significantly improves the performance of SkillNet on five modalities, on par with or even better than baselines with modality-specific pretraining. On the task of Chinese text-to-image retrieval, our final system achieves higher accuracy than existing leading systems including Wukong$_\text{ViT-B}$ and Wenlan 2.0 while using less number of activated parameters.
\end{abstract}

\section{Introduction}

In recent years, Transformer \cite{vaswani2017attention} and Transformer-based pretrained models \cite{devlin2018bert,radford2019language} have revolutionized natural language processing \cite{manning2022human} and there have been growing interests in extending the successful paradigm to broader artificial intelligence areas including computer vision \cite{bao2021beit,he2021masked,luo2021clip4clip}, speech processing \cite{baevski2020wav2vec} and program analysis \cite{feng2020codebert}.
Researchers from different communities have no communication barrier and typically repeat the same process: pretraining for each modality and finetuning all the model parameters for each task. 

Despite the remarkable progress made in artificial intelligence, existing methods differ from human learning in the following three aspects \cite{dean-pathways-2021}.
First, we human perceive the world using multiple senses. We know that the word ``dog'', the bark of a dog and the image/video of a dog all refer to the same concept. 
However, most existing methods only process one modality of information. Second, the human brain has around 100 billion neurons, of which different parts are specialized for different skills. 
When we accomplish a task, we only call upon a small fraction of neurons that are relevant to the task.
However, most existing methods activate all the model parameters. 
Third, when we solve a new problem or learn a new skill, we don't learn from nothing but combine old skills to learn new things quickly. 
However, existing methods typically learn for each task from scratch (or from a general or foundation model), resulting in hundreds of models for hundreds of tasks.

\begin{figure}[!t]
	\centering
	\subfigtopskip=0.5pt
	\subfigure[Fully activated dense model]{
	\includegraphics[width=0.42\linewidth]{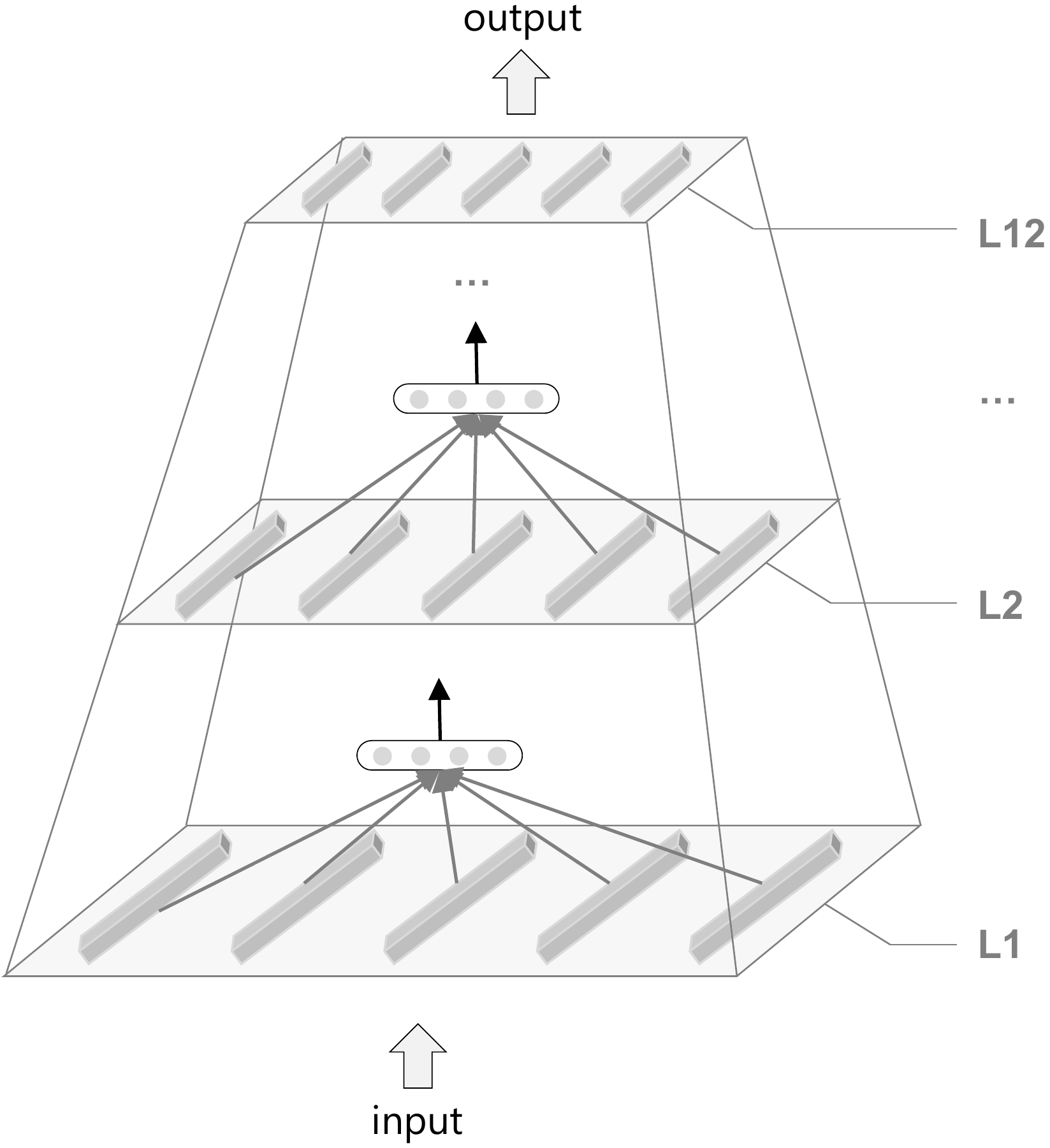}}\ \ \ \ \ \ \ \ \ \ \ \ 
	\subfigure[Sparsely activated MoE]{
	\includegraphics[width=0.42\linewidth]{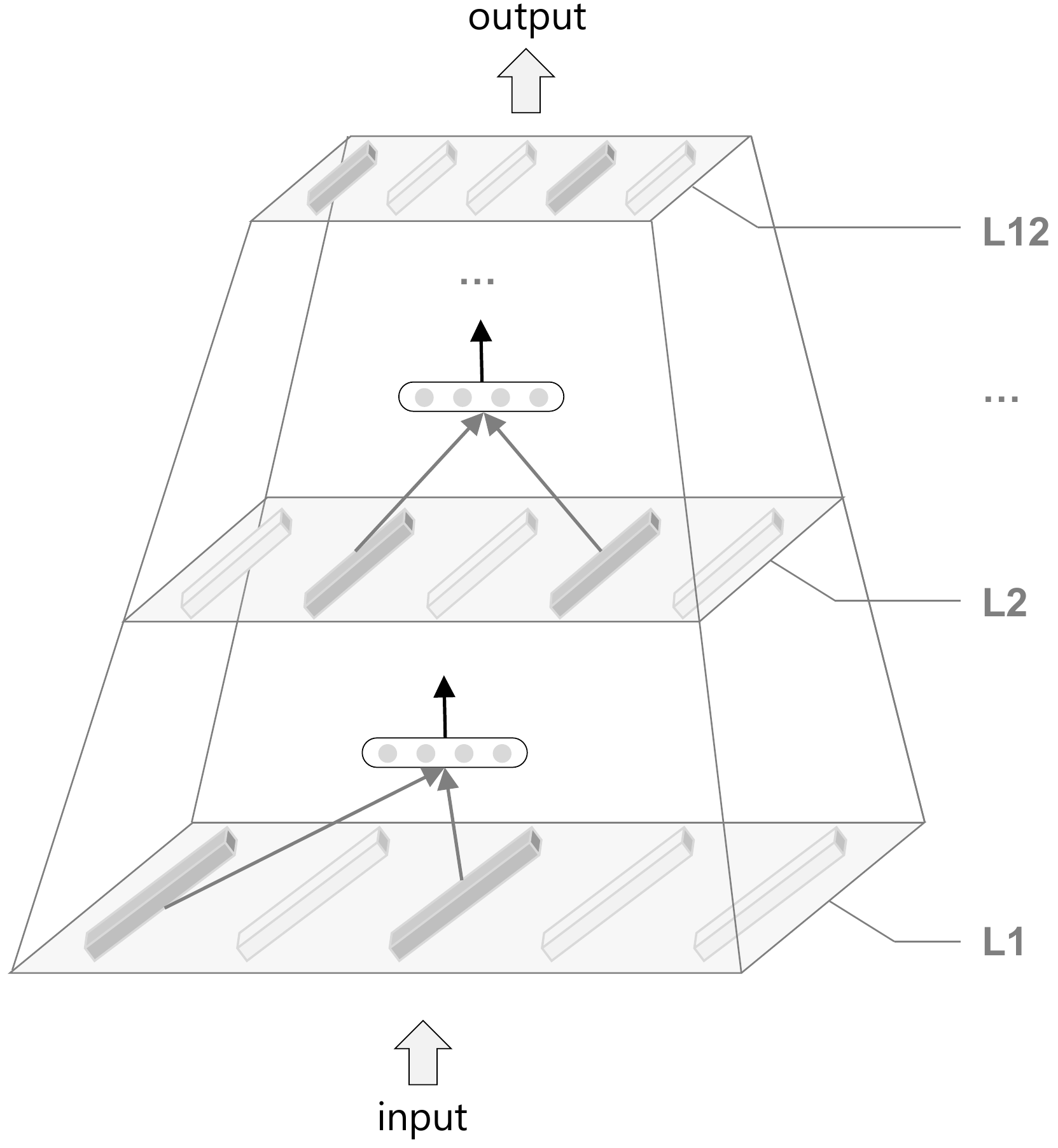}}
	
	\subfigure[SkillNet for ASR.]{
	\includegraphics[width=0.42\linewidth]{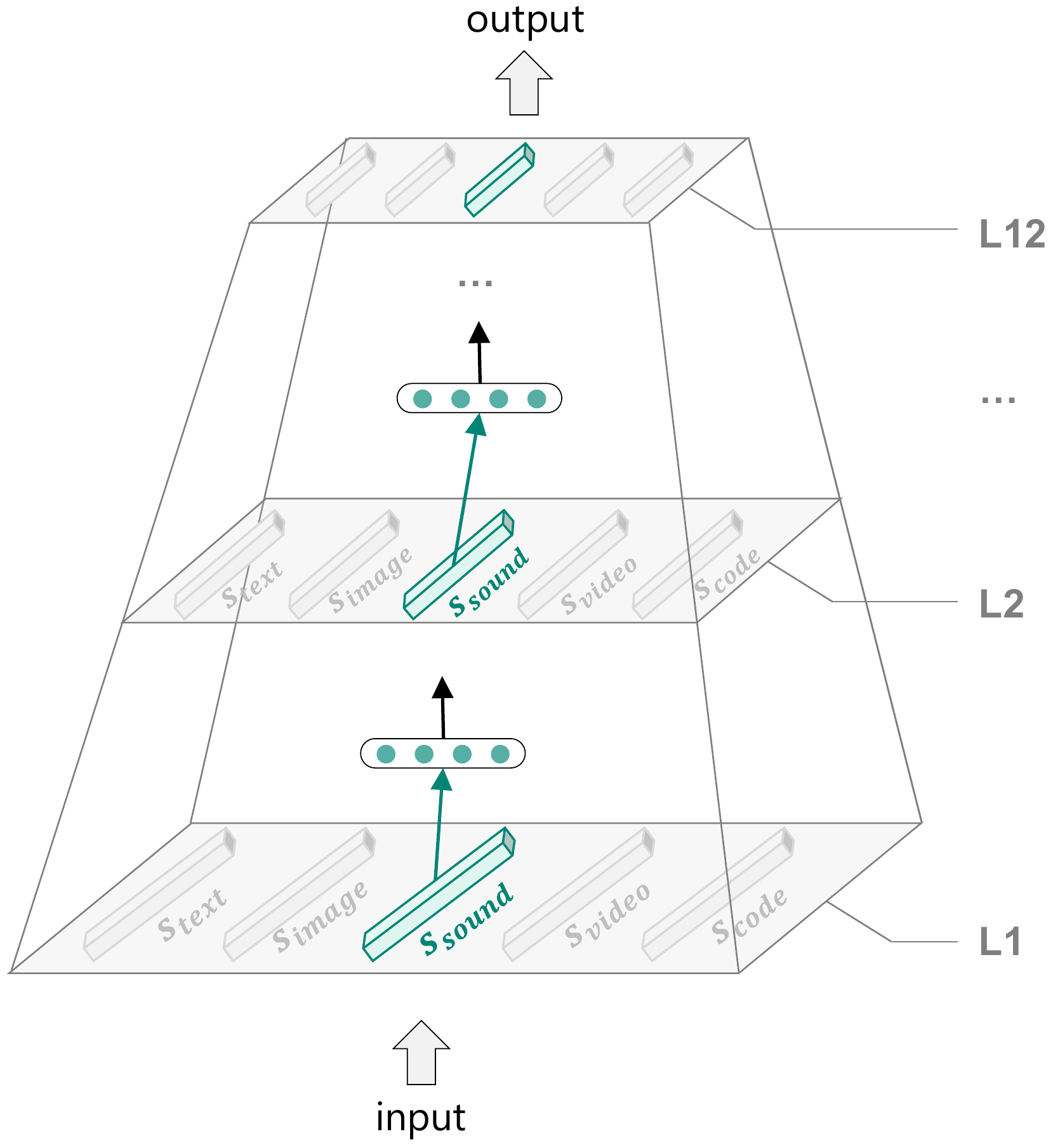}}\ \ \ \ \ \ \ \ \ \ \ \ 
	\subfigure[SkillNet for text-to-image retrieval]{
	\includegraphics[width=0.42\linewidth]{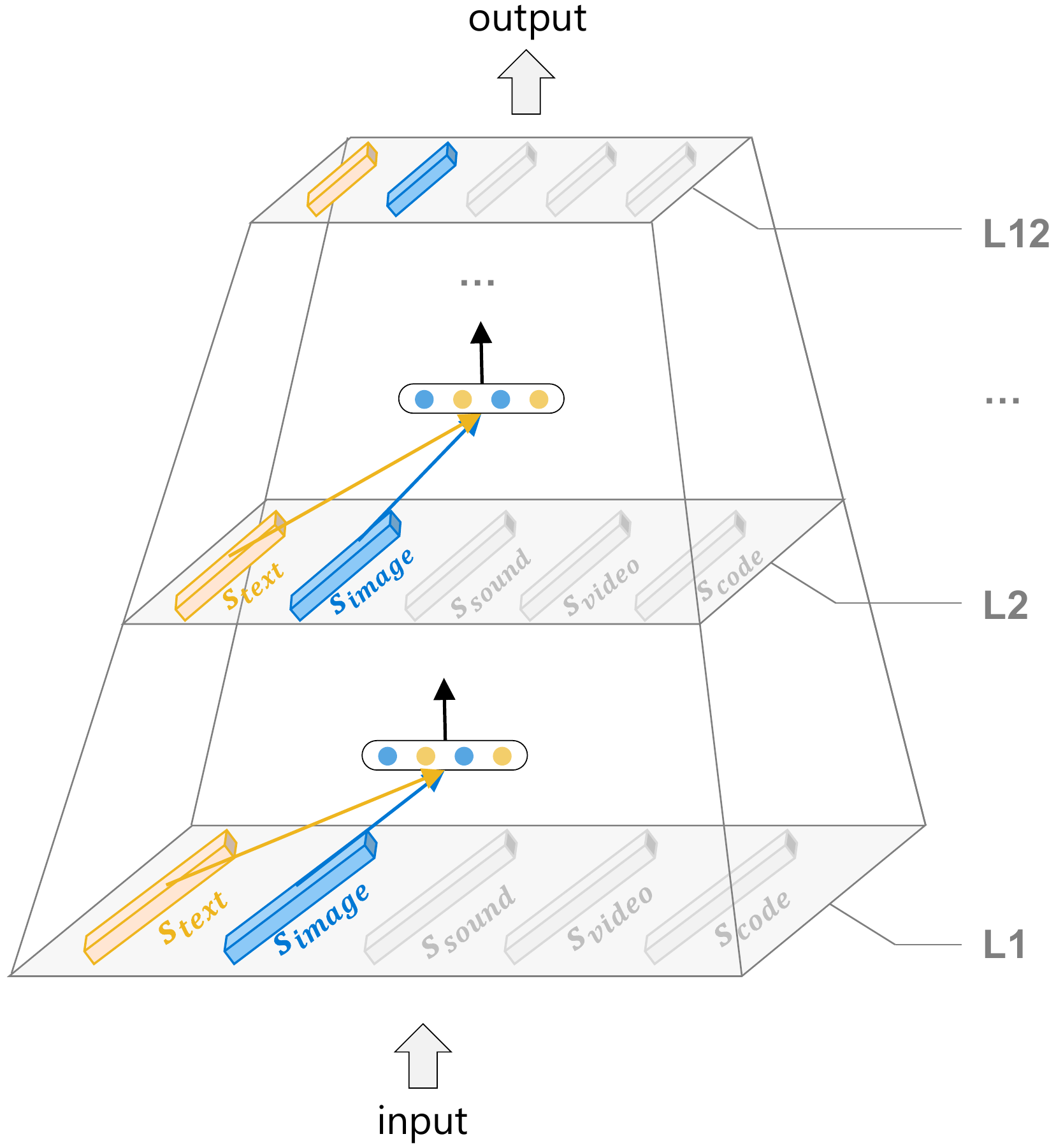}}
	
	\subfigure[SkillNet for text-to-video retrieval]{
	\includegraphics[width=0.42\linewidth]{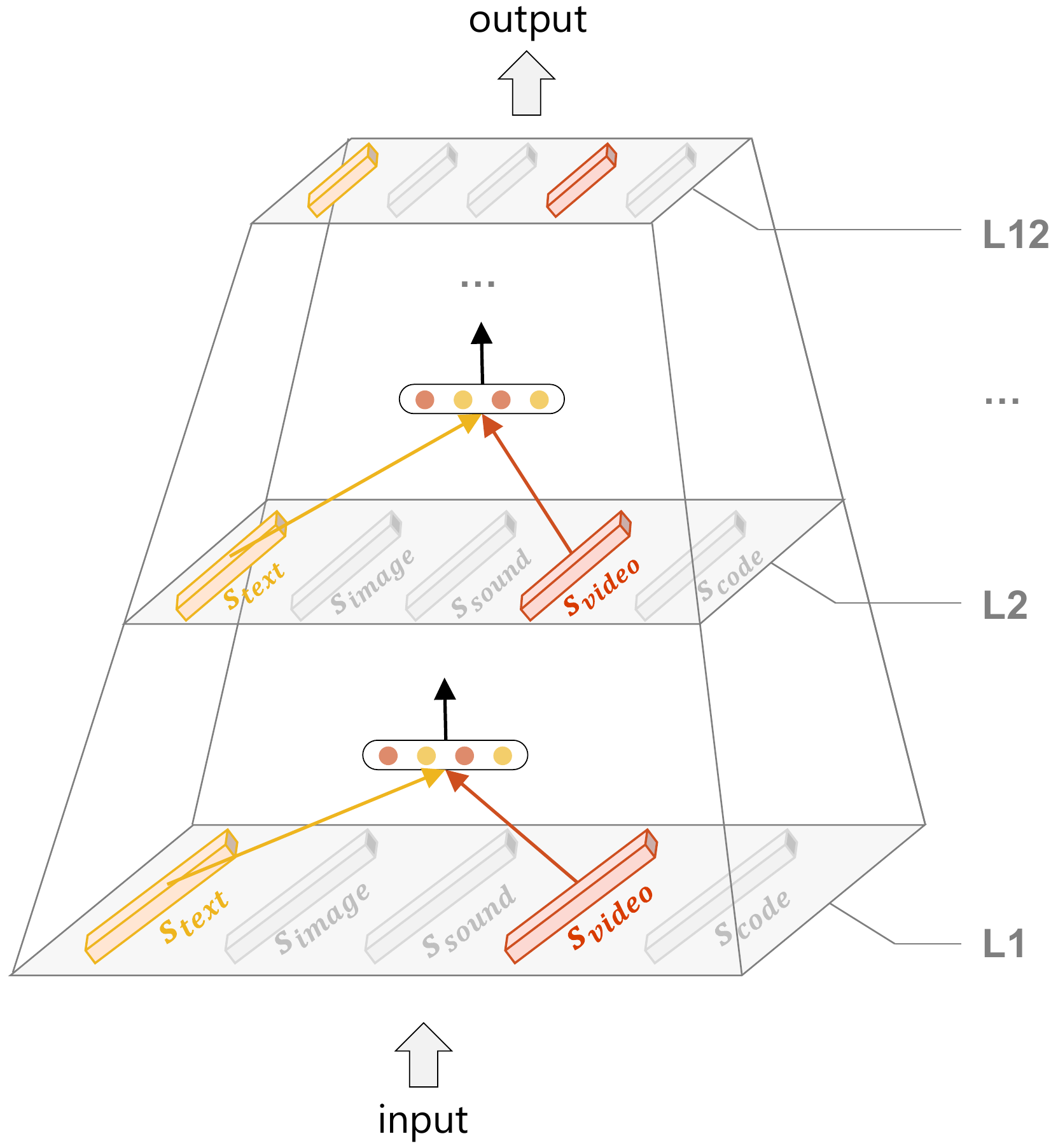}}\ \ \ \ \ \ \ \ \ \ \ \ 
	\subfigure[SkillNet for text-to-code retrieval]{
	\includegraphics[width=0.42\linewidth]{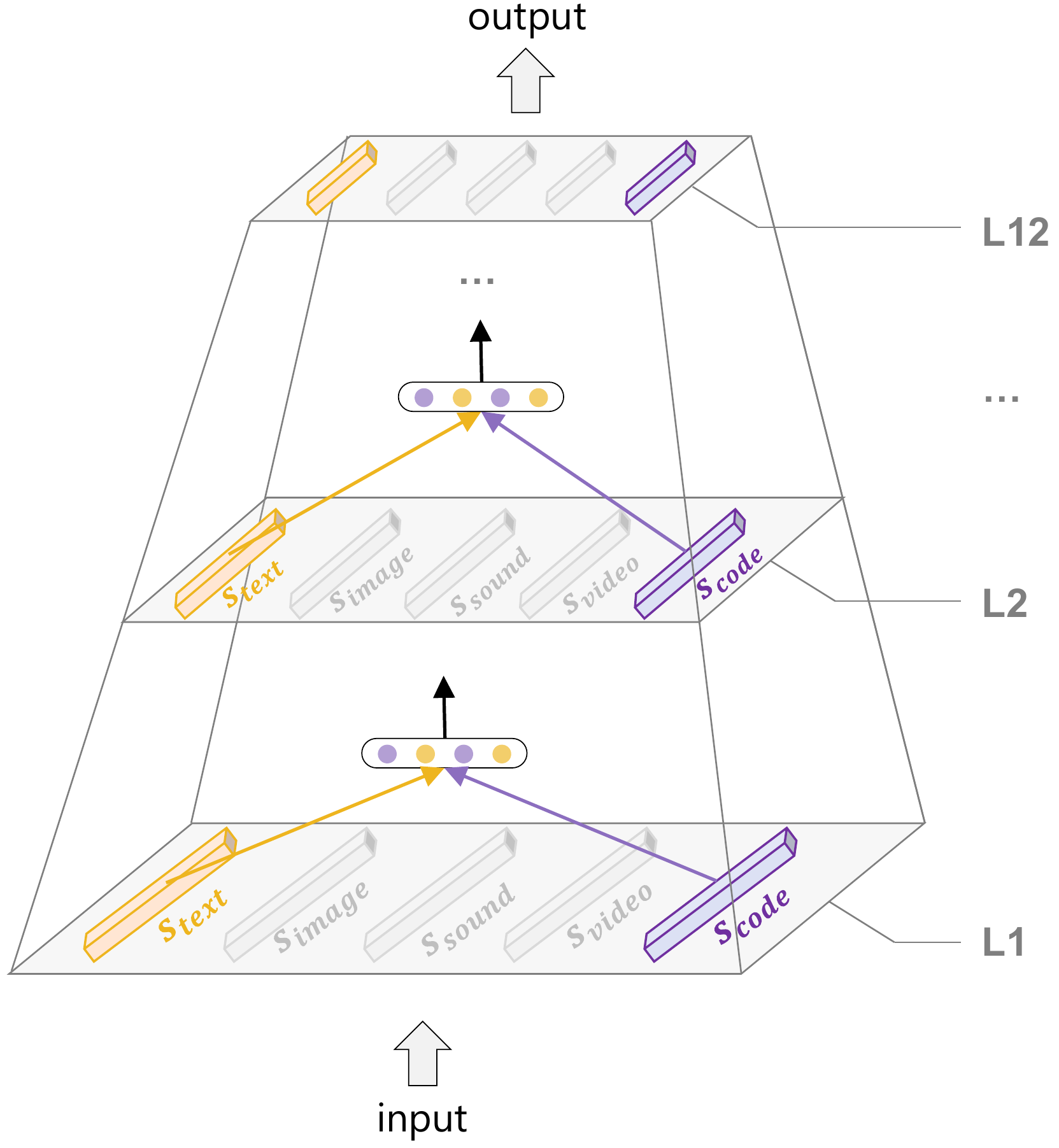}}
	\caption{ In SkillNet, each pillar refers to a skill. Pillars filled in color (e.g., yellow, blue, green, purple and red)  are activated. }
	\label{fig:model-intro}
\end{figure}

In this work, we propose a multitask multimodal approach called SkillNet.
We use a single model to handle multiple tasks that require the understanding of different modalities of information.
In SkillNet, different parts of the parameters are specialized for different skills.
When the model is applied to a downstream task, unlike traditional ``dense'' models that always activate all the model parameters, it ``sparsely'' activates parts of the parameters whose skills are relevant to the target task.
For example, we could define five modality-related skills \{$s_{text}$, $s_{image}$, $s_{sound}$, $s_{video}$, $s_{code}$\},
which are specialized for understanding text, image, sound, video and code, respectively.
Consider the task of automatic speech recognition (ASR), which only relates to the skill of auditory understanding (i.e., $s_{sound}$).
When SkillNet is applied to ASR, model parameters related to other four skills (i.e., \{$s_{text}$, $s_{image}$, $s_{video}$, $s_{code}$\}) are deactivated.
Similarly, for text-to-image retrieval, which is to find semantically related images given texts, only $s_{text}$ and $s_{image}$ are activated. Figure \ref{fig:model-intro} gives high-level illustrations of the aforementioned situations. 
There are many different ways to implement SkillNet. In this work, we provide a simple implementation on top of Transformer \cite{vaswani2017attention}.
Instead of producing general $K/Q/V$ vectors for each token, we activate different modality-specific parameters to produce different modality-specific $K/Q/V$ vectors before conducting multi-head attention.
The intuition is that we expect the model to call upon different parts as needed to process different types of signals and combine information from multiple senses to form our understanding about a concept (like the aforementioned example about the concept of dog).

We conduct experiments on tasks of five modalities, including text classification, automatic speech recognition, text-to-image retrieval, text-to-video retrieval and text-to-code retrieval. 
Results show that, SkillNet performs comparably to five modality-specific models with only one model file. 
Furthermore, after being pretrained, SkillNet performs better than systems with modality-specific pretraining on three of five modalities. On the task of Chinese text-to-image retrieval, SkillNet obtains higher accuracy than existing systems (e.g., Wukong$_\text{ViT-B}$ and Wenlan 2.0) while using less number of activated parameters.
Our work demonstrates the feasibility of developing one general model that is both accuracy and efficient to tackle multiple tasks of different modalities.

\section{Comparison to Existing Methods}
We describe the connections and differences of this work to related multimodal, multitask and mixture-of-experts methods.

\paragraph{Multimodal} Since there are large amounts of multimodal works, we only describe the closely related ones. Omnivore \cite{girdhar2022omnivore} uses a single model to process multiple visual modalities, including single-view 3D data, images and videos. 
VATT \cite{akbari2021vatt} learns multimodal representations on raw signals for video, audio and text. 
Compared to Omnivore and VATT, our work studies more modalities and our approach is sparse.
Data2vec \cite{baevski2022data2vec} is a general learning objective that manipulates over latent representations instead of modality-specific tokens. The same learning objective is used to learn for text, speech and vision. However, they don't perform multitask training. 
Our work is orthogonal to Data2vec and it is interesting to combine the advantages of Data2vec and SkillNet.

\paragraph{Multitask} This work also relates to multitask learning methods. Systems built upon Transformer typically use shared feature encoder plus task-specific prediction layers for understanding tasks \cite{liu2019multi} and use natural language prompts to steer encoder-decoder model for generation tasks \cite{raffel2019exploring}. Most existing multitask methods are dense \textemdash all the model parameters are activated. An exception is SkillNet-NLU and SkillNet-NLG \cite{tang2022skillnet,skillnetnlg}, recently introduced sparse models that perform multitask learning on text. This work can be viewed as an extension to the multimodal situation.

\paragraph{Mixture-of-Expert (MoE)} 
Transformer-based MoE methods typically include multiple homogeneous neural networks (called experts), which can be fully activated or partially activated guided by an additional gating function \cite{shazeer2017outrageously,lepikhin2020gshard,fedus2021switch,du2021glam}. 
However, it is unclear what type of knowledge is learned in each expert and why an expert is activated.
From this point of view, our approach can be viewed as a sparse multimodal MoE. Unlike traditional MoE methods, each expert in our model has a clear definition and the activation of each expert has a clear reason (judged by human experts).
For example, the expert corresponding to $s_{text}$ is responsible for understanding text signal and it is activated only if the input signal is text.

\section{Method}

This section gives our sparsely activated model SkillNet.
We first give a brief background on Transformer (\S \ref{sec:background}).
Then, we describe the model architecture of SkillNet (\S \ref{sec:model}).
Finally, we describe how to produce the embeddings for different modalities (\S \ref{sec:embedding}).

\subsection{Background on Transformer}\label{sec:background}
To make our paper self-contained, we briefly describe Transformer here. 
Transformer~\cite{vaswani2017attention} is a commonly used model architecture with multiple layers, each of which consists of a multi-head attention layer followed by a feed-forward network (FFN) layer.
The multi-head attention mechanism concatenates the output vectors of $H$ different heads and then linearly projects them by $W^O$:
\begin{equation}
\text{Multi-Head}(Q, K, V) = \text{Concat}(\text{head}_{1},...,\text{head}_{H}) W^{O},
\end{equation}
where $Q$ (Query), $K$ (Key), $V$ (Value) are the hidden representations of the previous layer.
\begin{wrapfigure}{r}{0.47\textwidth}
	\includegraphics[width=\linewidth]{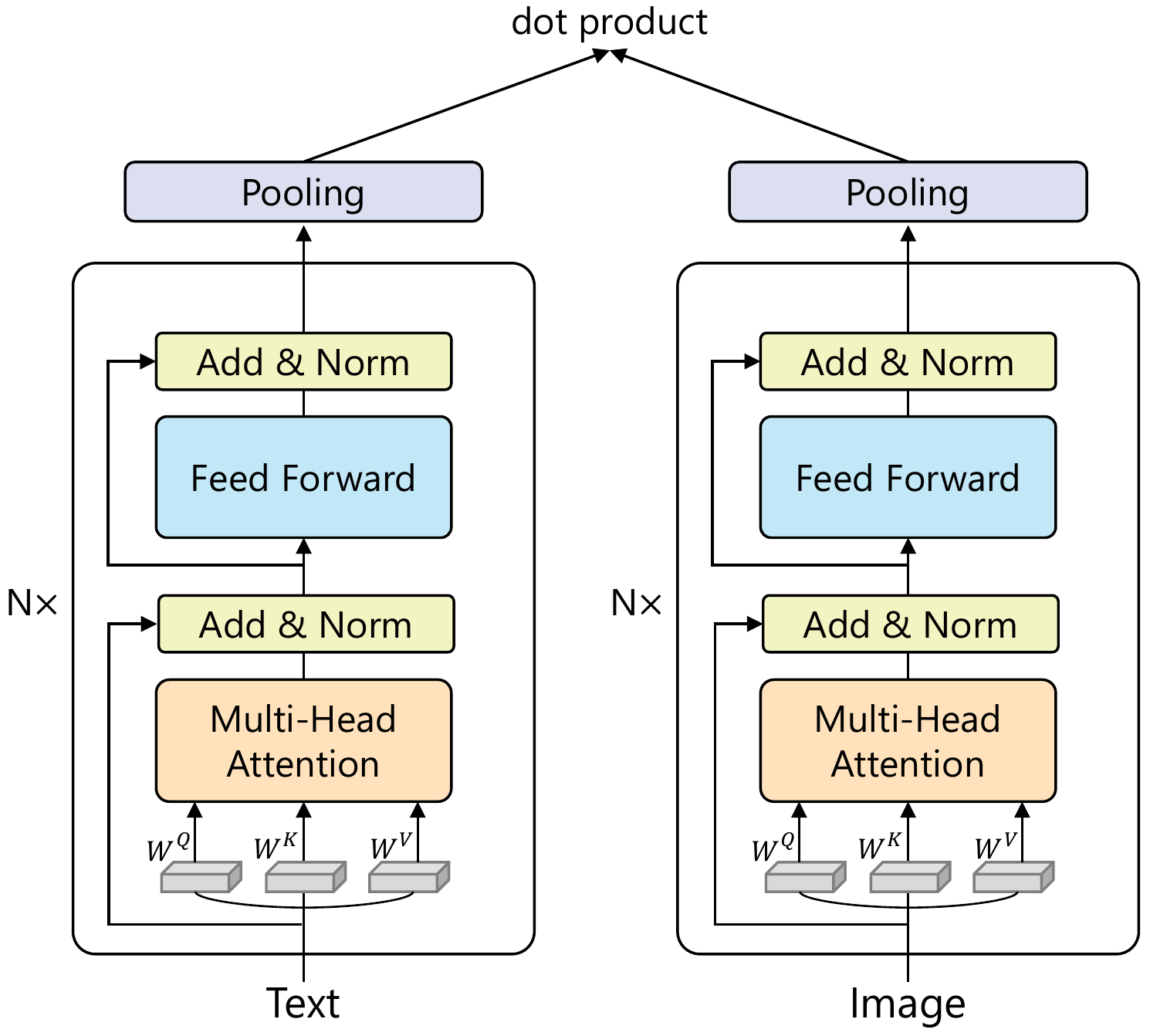}
	\caption{An illustration of image search with Transformer-based Siamese network. }
	\label{fig:transformer}
\end{wrapfigure}
In each head, $Q$, $K$ and $V$ are transformed with projection matrices before being fed to the attention function:
\begin{equation}
\text{head}_{i} = \text{Attention}(QW_i^Q, KW_i^K, VW_i^V),
\end{equation}
where $W_{i}^{Q}$, $W_{i}^{K}$ and $W_{i}^{V}$ are model parameters, and $i$ denotes the $i$-th head.
The attention function computes the dot products of the query with all keys, and uses softmax to obtain the weights on values:
\begin{equation} 
\text{Attention}(Q, K, V)=\text{softmax}(\frac{Q K^{T}}{\sqrt{d_{k}}}) V,
\end{equation} 
where $d_k$ is the dimension of key.
Finally, the FNN layer is applied to obtain the final representations.

Residual connection~\cite{he2016deep} and layer normalization~\cite{ba2016layer} are adopted for both multi-head attention layer and FFN layer. 
Since Transformer is prevalent, we exclude the details and refer readers to the original paper.

We use image search via Siamese network as the running example to show how to apply Transformer to downstream tasks. As shown in Figure \ref{fig:transformer}, 
 two Transformers are used to encode the text and the image, respectively.
For each side, we take the vector of the first token (\texttt{[CLS]}) to represent the input.
The semantic similarity between text and image is computed using dot product or cosine.

\subsection{Architecture of SkillNet}\label{sec:model}
We build our SkillNet model by using Transformer~\cite{vaswani2017attention} as the backbone.
Specifically, we modify the multi-head attention of each Transformer layer as follows.
Instead of producing general $K/Q/V$ vectors for each token, we activate different modality-specific parameters to produce different modality-specific $K/Q/V$ vectors before conducting multi-head attention.
Take $Q$ as an example. Instead of having only one projection matrix $W_i^{Q}$ for all queries, we have five projection parameter matrices \{$W_i^{Q_{text}}$, $W_i^{Q_{image}}$, $W_i^{Q_{sound}}$, $W_i^{Q_{video}}$, $W_i^{Q_{code}}$\}, of which each item stands for a skill of understanding the information of a particular modality.
When the model is applied to a task, we only activate the corresponding projection matrices of relevant skills. Similar modifications are made for keys and values.
The computation of a head is modified as follows.

\begin{equation}
\text{head}_{i}^{\text{skill}} = \text{Attention}(Q
W_{i}^{Q_{Activated}},
K
W_{i}^{K_{Activated}},V
W_{i}^{V_{Activated}})
\end{equation}
\begin{equation}
W_{i}^{Q_{Activated}} = 
\begin{cases}
W_{i}^{Q_{text}} & \text{if} \ \texttt{s}_{text} \ \text{is activated} \\
W_{i}^{Q_{image}} & \text{if} \ \texttt{s}_{image} \ \text{is activated} \\
W_{i}^{Q_{sound}} & \text{if} \ \texttt{s}_{sound} \ \text{is activated} \\
W_{i}^{Q_{video}} & \text{if} \ \texttt{s}_{video} \ \text{is activated} \\
W_{i}^{Q_{code}} & \text{if} \ \texttt{s}_{code} \ \text{is activated} \\
\end{cases}
\end{equation}

\begin{figure}[t]
	\centering
	\includegraphics[width=\textwidth]{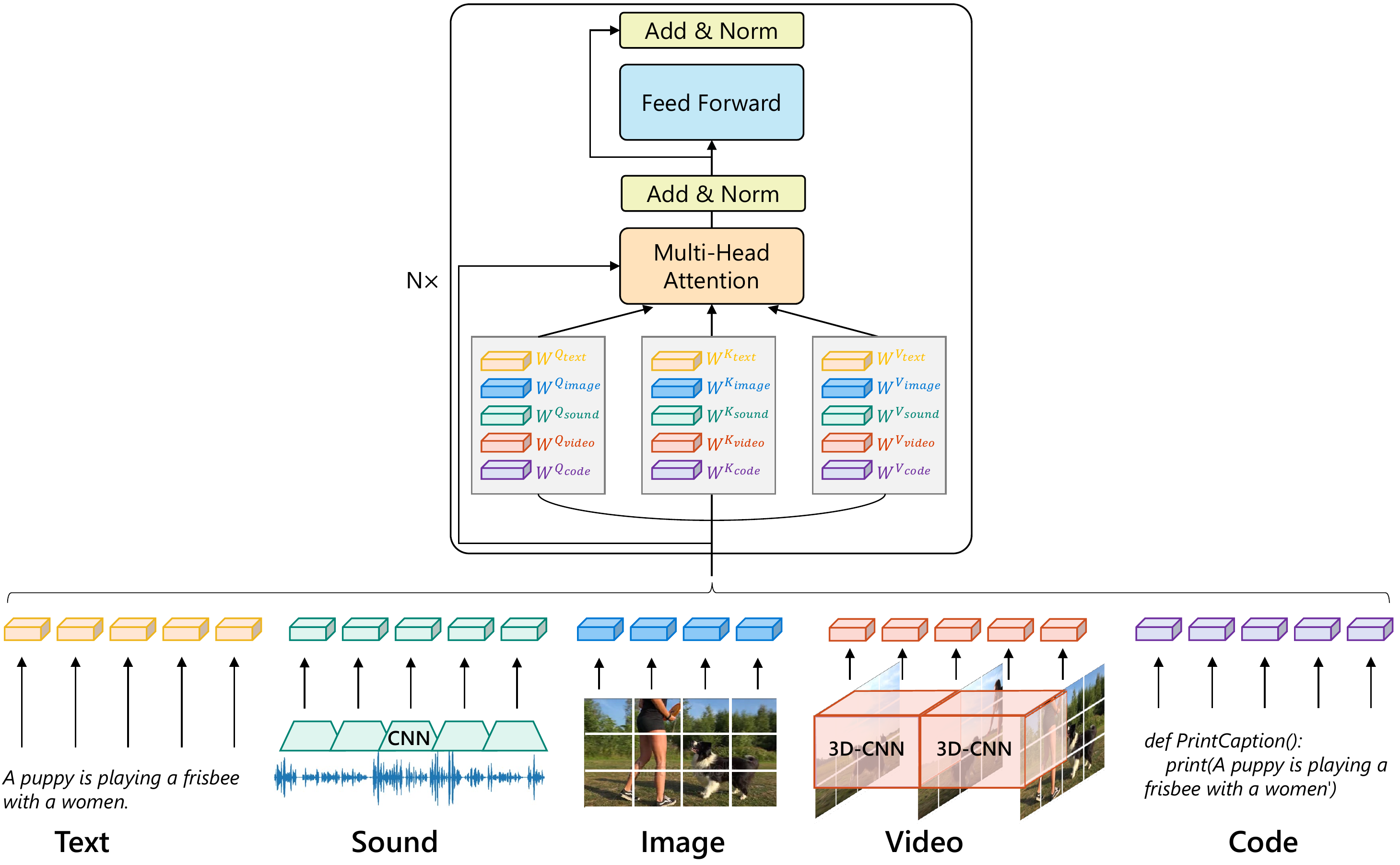}
	\caption{An illustration of the pipeline and the embeddings of different modalities.}
	\label{fig:model-embedding}
\end{figure}

As shown in Figure \ref{fig:model-embedding}, we only need one model to handle the the task of image retrieval, where we activate \texttt{s}$_{text}$ and \texttt{s}$_{image}$ for the text encode and image encoder, respectively.

\begin{figure}[t]
	\centering
	\includegraphics[width=\textwidth]{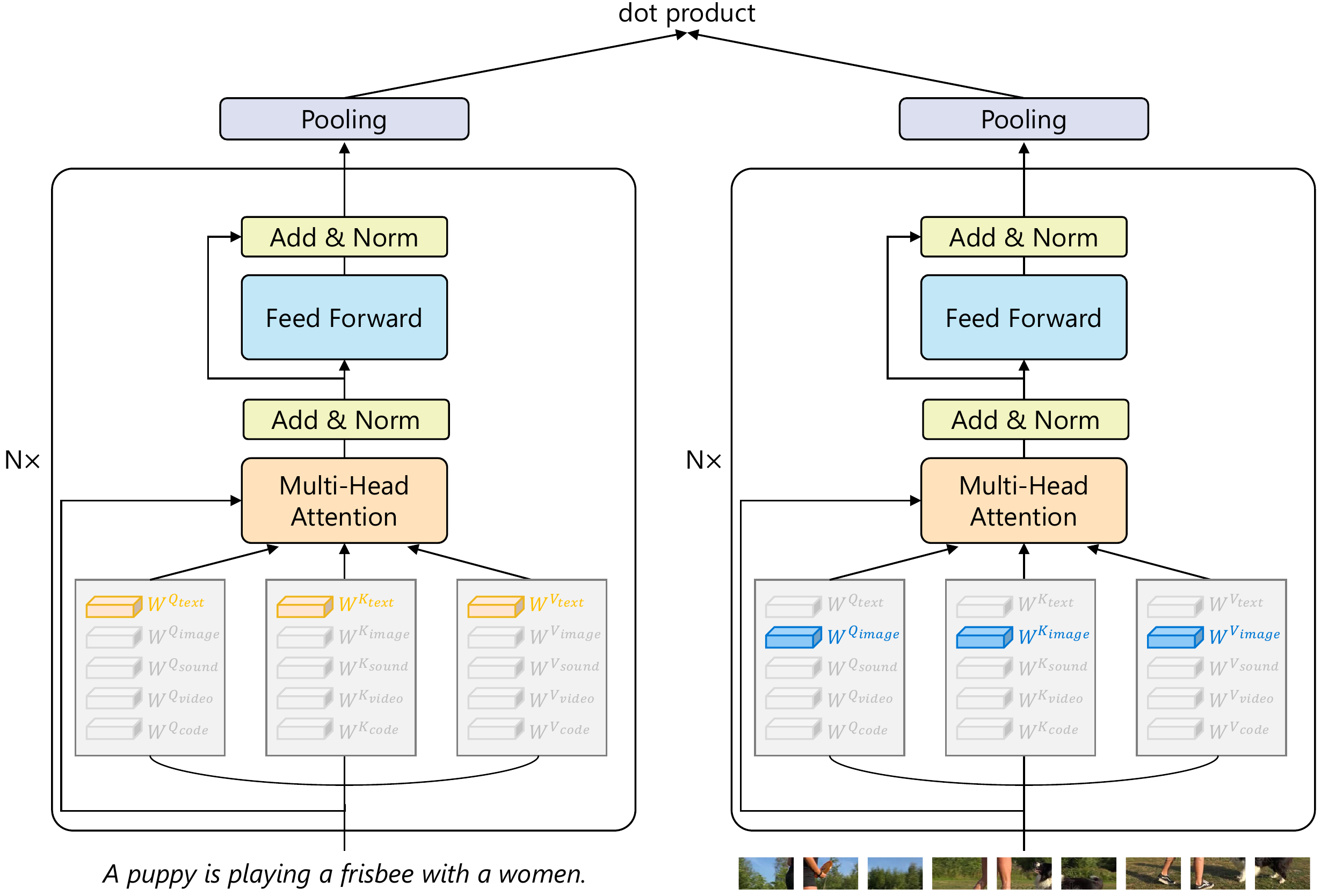}
	\caption{Architecture of SkillNet for image retrieval. Text encoder and image encoder are two pathways of one shared model \textemdash \ \texttt{s}$_{text}$ and \texttt{s}$_{image}$ are activated for the text encoder and the image encoder, respectively.}
	\label{fig:model-architecture}
\end{figure}

\subsection{Embeddings}\label{sec:embedding}
We describe how to produce the embeddings for different modalities.

\paragraph{Text} Following BERT \cite{devlin2018bert}, we tokenize a text into a sequence of wordpiece tokens \cite{yonghui2016bridging} and build the embedding of each wordpiece by adding up its token embedding, position embedding and segment embedding. We also add a special classification token \texttt{[CLS$_{\texttt{text}}$]} at the beginning of a sequence to produce the representation of the sequence.
If the input includes two segments, we add a special token \texttt{[SEP]} between the two segments.

\paragraph{Sound} Given a raw waveform as the input, we follow wav2vec \cite{baevski2020wav2vec} and use convolutional network to produce a sequence of vectors as the embeddings. Specifically, we use seven convolutions with 512 channels, strides of (5,2,2,2,2,2,2) and kernel widths of (10,3,3,3,3,2,2) to generate a vector
sequence from a 20ms framerate sampled at 16KHz. 
After that, we adopt a 1D convolutional network to transform the vector sequence to 768 dimensional embeddings, which are summed up with their corresponding position embeddings as the final sound embeddings.

\paragraph{Image} Following Vision Transformer (ViT)\cite{dosovitskiy2020image}, we build patch embeddings for each image. We first reshape each image of $\mathbf{x} \in \mathbb{R}^{H \times W \times C}$ into 2D patches of $\mathbf{x}_p \in \mathbb{R}^{N \times (P^2\cdot  C)}$, where $(H,W)$ is the image resolution, $(P,P)$ is the resolution of each patch, $N$ is the number of patches and $C$ is the number of image channels (e.g. 3 for RGB). Then, a 2D convolutional network is applied to transform patch pixels to 768 dimensional embeddings, which are added with the corresponding position embeddings as the final patch embeddings.\footnote{In this work, we use different positional embeddings for different modalities.} We add a special token \texttt{[CLS$_{\texttt{image}}$]} at the beginning of each sequence to produce the representation of the image.

\paragraph{Video} We follow Vivit \cite{arnab2021vivit}, an extension of ViT for video, to produce video embeddings. Given a video $V \in R^{T \times H \times W \times C}$, where $T$ is the number of sampled frames, we extract $[T/t]\cdot[H/h]\cdot[W/w]$ non-overlapping, spatio-temporal ``tubes'' and use a 3D convolution to produce a representation for each tube. We further add $[T/t] + [H/h] + [W/w]$ positional embeddings and concatenate a special token \texttt{[CLS$_{\texttt{video}}$]}  at the beginning of each sequence to represent the whole video input.

\paragraph{Code} We follow CodeBERT \cite{feng2020codebert} to produce code embeddings. We tokenize a code snippet to a sequence of code-specific wordpiece tokens. The final embedding of each token is the sum of token embedding, position embedding and segment embedding. A special token \texttt{[CLS$_{\texttt{code}}$]} is added to the beginning of each sequence to produce the embedding of the entire code.

\section{Tasks}
In this section, we first describe downstream tasks involving five modalities in \S \ref{sec:downstream-tasks}. 
Each modality relates to an active research area that covers many tasks. We select one task for each modality with preferences for well recognized tasks (e.g., ASR) and tasks relate to multiple modalities (e.g., video/code retrieval).
Since our framework also supports sparsely activated pretraining, we conduct multimodal pretraining to initialize the model parameters. The pretraining tasks are described in \S \ref{sec:pre-training}.

\subsection{Downstream Tasks}\label{sec:downstream-tasks}

\begin{table*}[h]
\centering
\begin{tabular}{ccccccc}
\toprule
\multirow{2}{*}{\bf Task Id} & \multirow{2}{*}{\bf Task} & \multicolumn{5}{c}{\bf Skills} \\
\cmidrule(lr){3-7}
& & \texttt{s}$_{text}$ & \texttt{s}$_{image}$ & \texttt{s}$_{sound}$ & \texttt{s}$_{video}$ & \texttt{s}$_{code}$  \\
\midrule
\texttt{T1} & Text Classification & \checkmark & & & &  \\
\texttt{T2} & Automatic Speech Recognition & & & \checkmark & &  \\
\texttt{T3} & Text-to-Image Retrieval & \checkmark & \checkmark & & &  \\
\texttt{T4} & Text-to-Video Retrieval & \checkmark & & & \checkmark &  \\
\texttt{T5} & Text-to-Code Retrieval & \checkmark & & & & \checkmark  \\
\bottomrule
\end{tabular}
\caption{Relations between tasks and skills. Relevant skills for each task are marked with ticks.}
\label{tab:task-skills}
\end{table*}

\paragraph{Text}
Text classification is a classic and fundamental text understanding task \cite{minaee2021deep}.
Given a sentence as the input, the task is to predict which category the sentence belongs to.
Following BERT \cite{devlin2018bert}, we add a \texttt{[CLS$_{\texttt{text}}$]} token at the beginning of each sentence to represent the meaning of the whole sentence.
For the task of text classification, only the parameters that relate to \texttt{s}$_{text}$ are activated.

\paragraph{Sound}
Automatic speech recognition (ASR) is to convert speech to text \cite{hinton2012deep}.
Following wave2vec \cite{baevski2020wav2vec}, we produce speech features and generate a transcription by performing token-level classification.
Connectionist temporal classification loss \cite{graves2006connectionist} is adopted for model training.
For the task of ASR, only the parameters that relate to \texttt{s}$_{sound}$ are activated.

\paragraph{Image} 
We consider text-to-image retrieval.
Given a text as the query, the task is to find the target image from a set of candidates.
Considering the efficiency of the inference stage, we use two separate passes (like Siamese Network) to produce text and image vectors separately with no cross-modality attention. 
Notably, we use the same model with different activation configurations (i.e., \texttt{s}$_{text}$ is activated for text and the \texttt{s}$_{image}$ is activated for image) to produce text and image vectors.
The semantic similarity between a text and an image is calculated with dot product or cosine function.

\paragraph{Video} 
We consider text-to-video retrieval.
Given a text as the query, the task is to find the target video from a set of candidates. 
The framework is similar to the aforementioned image retrieval. 
We use the same model with different activated parameters (i.e., \texttt{s}$_{text}$ is avtivated for text and \texttt{s}$_{video}$ is activated for video) to produce text and video vectors separately. 

\paragraph{Code} 
We consider natural language code retrieval.
Given a text as the query, the task is to find the most relevant code from a set of candidates.
We use the same model with different activated parameters (i.e., \texttt{s}$_{text}$ for text and \texttt{s}$_{code}$ for code) to produce text and code vectors separately. The framework is similar to image retrieval.

\subsection{Pretraining Tasks}\label{sec:pre-training}
Recap that our approach also supports multimodal pretraining with sparse activation.
We describe the pretraining tasks for each modality here.

\paragraph{Text}
We adopt masked language modeling (MLM) as the pre-training task \cite{devlin2018bert,liu2019roberta}.
Given a text, we randomly mask 15\% of the tokens. Each masked token is replaced with a special [MASK] token 80\% of the time, a random token 10\% of the time, and left unchanged for the remaining 10\% of the time.

\paragraph{Sound} 
We develop a simplified version of HuBERT \cite{hsu2021hubert} and pretrain through predicting the categories of the masked sound tokens, whose target labels are produced with an offline clustering process.
We set the number of clusters to 100 and use k-mean clustering with Mel-Frequency Cepstral Coefficients (MFCCs) acoustic features.
We use the same masking strategies of wav2vec2 \cite{baevski2020wav2vec}, where about 5\% of the time-steps are randomly sampled as start indices and the subsequent 10 time-steps are masked.

\paragraph{Image} We follow CLIP \cite{radford2021learning}  and use contrastive objectives for pretraining. We use the same architecture for image retrieval as illustrated in  \S \ref{sec:downstream-tasks} and adopt in-batch negative sampling. 

\paragraph{Video} Similar to the configuration of image pretraining, we consider a contrastive pretraining task of text-video matching. In-batch negative sampling is adopted.

\paragraph{Code}
Like CodeBERT \cite{feng2020codebert}, we concatenate code and text, separate them with \texttt{[SEP]} and randomly mask 15\% of the tokens. The pretraining task is to predict the masked tokens. 

\section{Experiments}


\subsection{Setup}

We compare to the following baselines. 

    $\bullet$ \textbf{Modality-specific models}. We train five different models for different modalities. The model architecture for each modality is the standard Transformer.
    
    $\bullet$ \textbf{Dense multimodal baseline}. We train a multimodal model that jointly learns for five modalities. This is a dense model in that all these modalities share a common standard Transformer architecture, which is equivalant to SkillNet with only one skill and that skill is always activated.
    
    $\bullet$ \textbf{MoE multimodal baseline}. We train a Mixture-of-Expert (MoE) \cite{lepikhin2020gshard} baseline and set the number of experts as the number of skills of SkillNet (i.e., 5). There is a gating function to selectively active top-2 experts for each token.
    
We implement SkillNet on top of  HuggingFace’s Transformers~\cite{wolf2020huggingfaces}. We conduct experiment with 12 Transformer encoder layers and 768 hidden state dimensions and leave the extension to larger model scales to the future.
Since the parameters of SkillNet can be pretrained (as described in \S \ref{sec:pre-training}), we have two model configurations, depending on whether the parameters are pretrained in the same sparsely activated manner.
We also compare to baselines with modality-specific pretraining. For text, we compare to \cite{tang2022skillnet}, which uses the superset of our text pretraining corpus to pretrain BERT. 
For image, we compare to Wukong$_\text{ViT-B}$ \cite{gu2022wukong}, which has the similar model scale (with 12 Transformer layers) and is pretrained with a superset of our image pretraining data. 
For speech, video and code, we pretrain modality-specific models with the same amount of pretraining data of SkillNet. 

Details about the datasets and training process are given in the Appendix.

\subsection{Results and Analysis}

Table \ref{tab:results} gives the results on five tasks.  
\begin{table*}[!t]
\centering
\small
\begin{tabular}{l|ccccc}
\toprule
Method &  \texttt{Text} & \texttt{Image} & \texttt{Sound} & \texttt{Video} & \texttt{Code}  \\
\midrule
Modality-specific models & 0.48 &69.63 &0.20 &63.18 &53.97 \\ 
Dense multimodal baseline   &0.48 &55.70 &0.23 &19.46 &57.59 \\
MoE multimodal baseline    & 0.49&60.93 &0.19 &64.81 &50.04 \\
{SkillNet w/o pretraining}  & 0.48 &68.76 &0.20 &66.49 &60.14  \\
\midrule
Baselines with modality-specific pretraining & \ \ 0.56$^*$ & \ \  71.70$^\dagger$& 0.17 & 77.31 & 66.33 \\
\textbf{SkillNet}  & \textbf{0.57}& \textbf{73.59}&\textbf{0.17}& \textbf{81.77}&\textbf{70.66} \\
\bottomrule
\end{tabular}
\caption{Results on five tasks. Evaluation metrics for five modalities are accuracy, Recall@10, CER (lower is better), Recall@10 and Recall@10, respectively. The result tagged with $^*$ is from \cite{tang2022skillnet}, whose pretraining text corpus is the superset of our work. The result tagged with $^\dagger$ is from the previous best system for Chinese text-to-image retrieval \cite{gu2022wukong}, whose pretraining image corpus is also the superset of our image pretraining data. }
\label{tab:results}
\end{table*}
Systems in the first group are not pretrained. We can see that SkillNet performs comparably to modality-specific models. 
An interesting finding is that the joint model with a dense encoder is not friendly to the low-resource task like text-to-video, but this phenomenon does not exist in either MoE system or SkillNet.
The second group includes systems with modality-specific pretraining or joint pretraining. We can see that pretraining consistently improves the performance of SkillNet on all five tasks, even better than modality-specific pretraining on image, video and code.

On the task of text-to-image retrieval, SkillNet achieves better accuracy compared to existing leading systems but using less number of activated parameters. 
Numbers are given in Table \ref{tab:activate-params}.
Since the parameters of Wenlan 2.0 and Wukong$_\text{ViT-B}$ are not reported in their papers, we calculate their parameters based on their model descriptions. 
The parameters of Wenlan 2.0 \cite{fei2021wenlan} include three parts, an image encoder consisting of an EfficientNet-B7 \cite{tan2019efficientnet} (66M) and four Transformer encoder layers (50M), a text encoder RoBERTa-Large \cite{cui-etal-2020-revisiting} (326M) and a cross-modal projection layer with two fully-connected layers (3M).  Wukong$_\text{ViT-B}$ \cite{gu2022wukong} includes a Vision Transformer (ViT) \cite{dosovitskiy2021an} (86M) as the image encoder, a standard decoder-only transformer (110M) as  the text encoder and a linear cross-modal projection layer (0.6M). 
\begin{table*}[h]
\centering
\begin{tabular}{l|c|cc}
\toprule
Method &  Number of Activated Params & R@1 & R@10\\
\midrule
Wenlan 2.0 \cite{fei2021wenlan}& 445M & 34.1&69.1\\
Wukong$_\text{ViT-B}$ \cite{gu2022wukong} & 197M &  36.7& 71.7\\
SkillNet &124M  & 37.0& 73.6\\
\bottomrule
\end{tabular}
\caption{Performance and activated model parameters of text-to-image retrieval methods.}
\label{tab:activate-params}
\end{table*}

\begin{figure}[!h]
	\centering
	\subfigtopskip=0.5pt
	\subfigure[SkillNet for text-to-image retrieval.]{
	\includegraphics[width=0.48\linewidth]{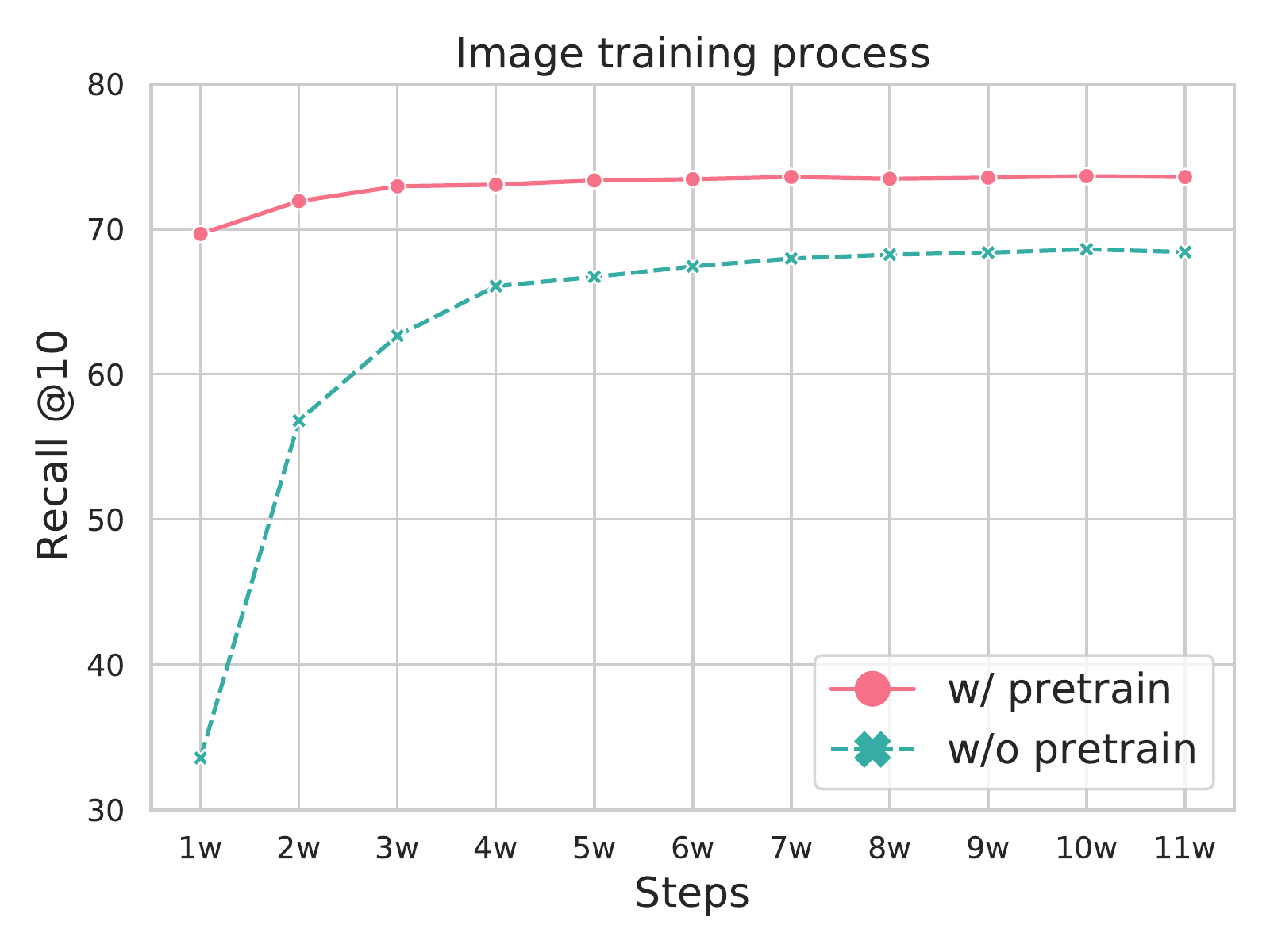}}\ \ \ \ 
	\subfigure[SkillNet for text-to-video retrieval.]{
	\includegraphics[width=0.48\linewidth]{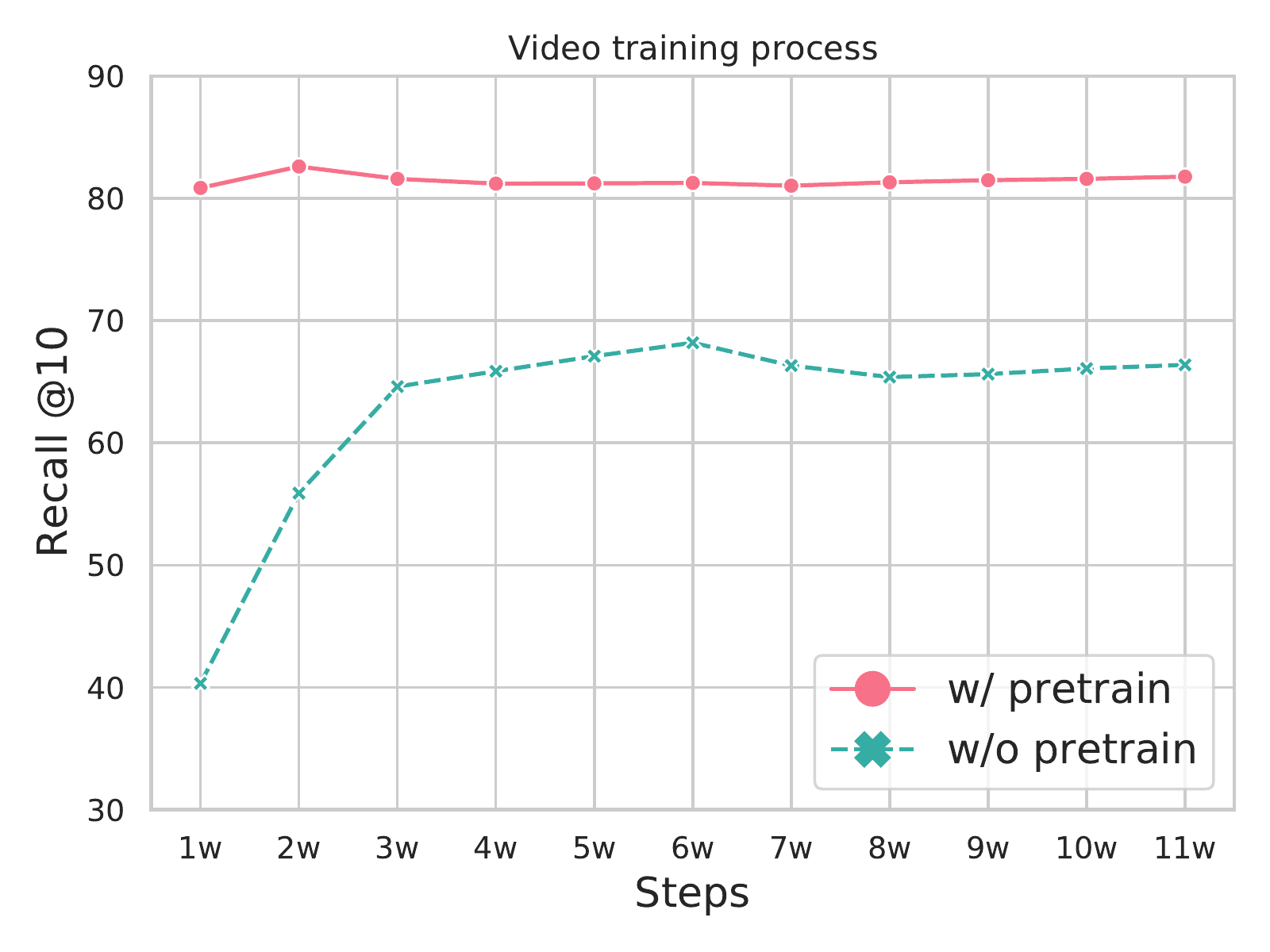}}
	
	\subfigure[SkillNet for ASR.]{
	\includegraphics[width=0.48\linewidth]{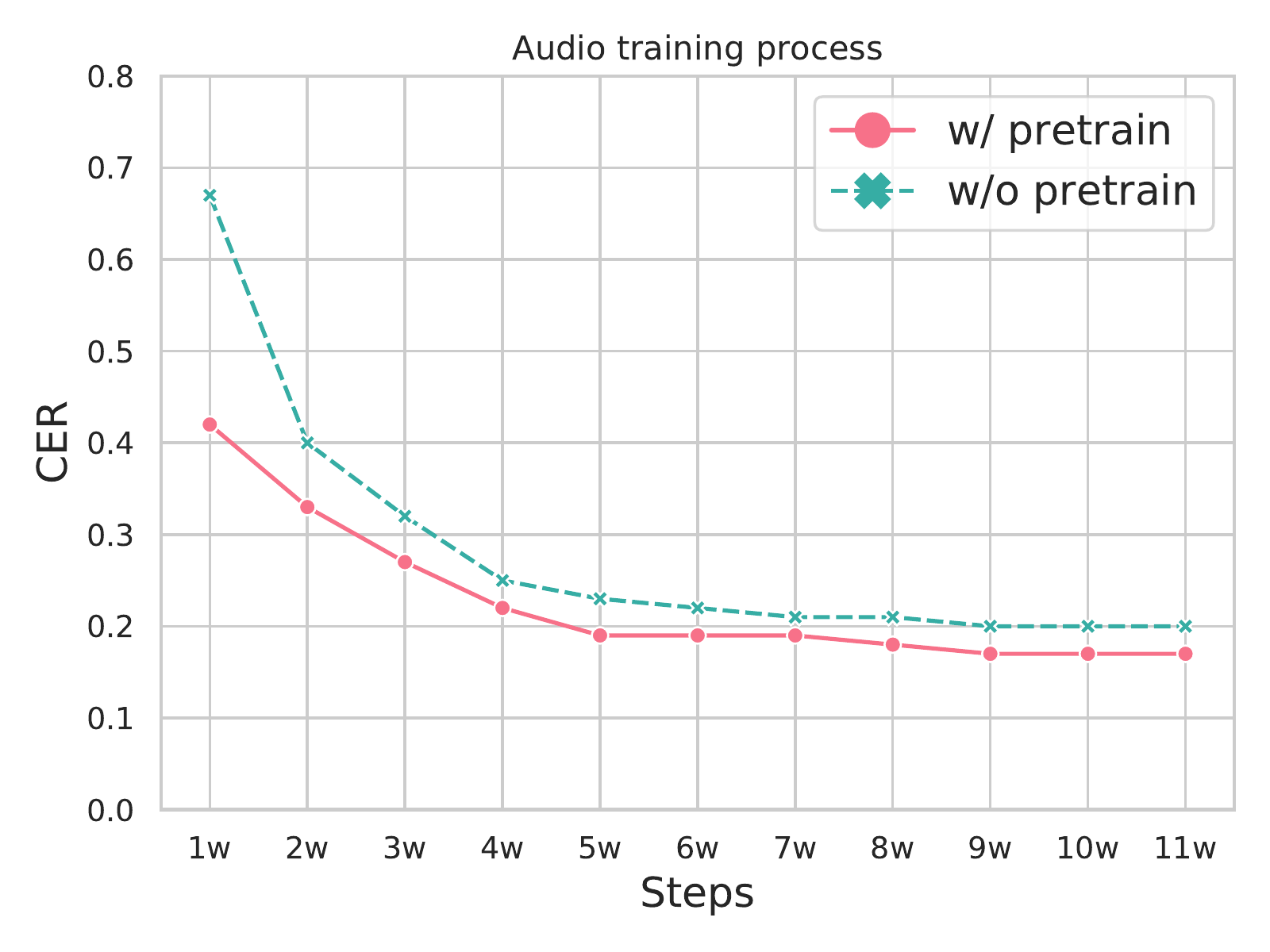}}\ \ \ \ 
	\subfigure[SkillNet for text-to-code retrieval.]{
	\includegraphics[width=0.48\linewidth]{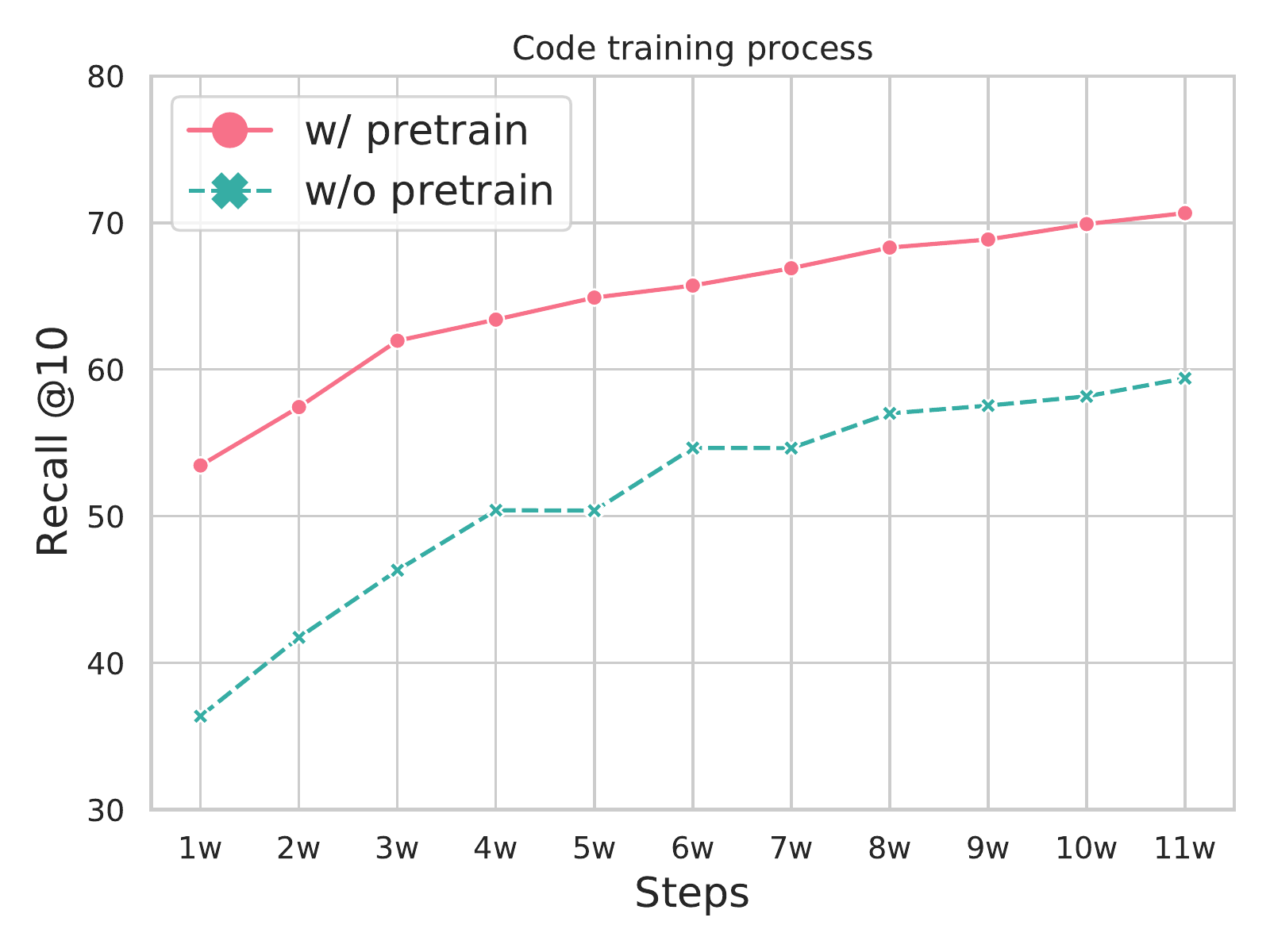}}
	\caption{Performance of SkillNet with different finetuning steps. X-axis stands for the  training steps. Y-axis stands for the evaluation metric (lower is better for CER). }
	\label{fig:curve}
\end{figure}

Figure \ref{fig:curve} shows the learning curves of SkillNet with or without pretraining on different tasks.
We can see that in general pretraining gives the model a good starting point and leads to better accuracy.

\begin{figure}[h]
	\centering
	\includegraphics[width=\textwidth]{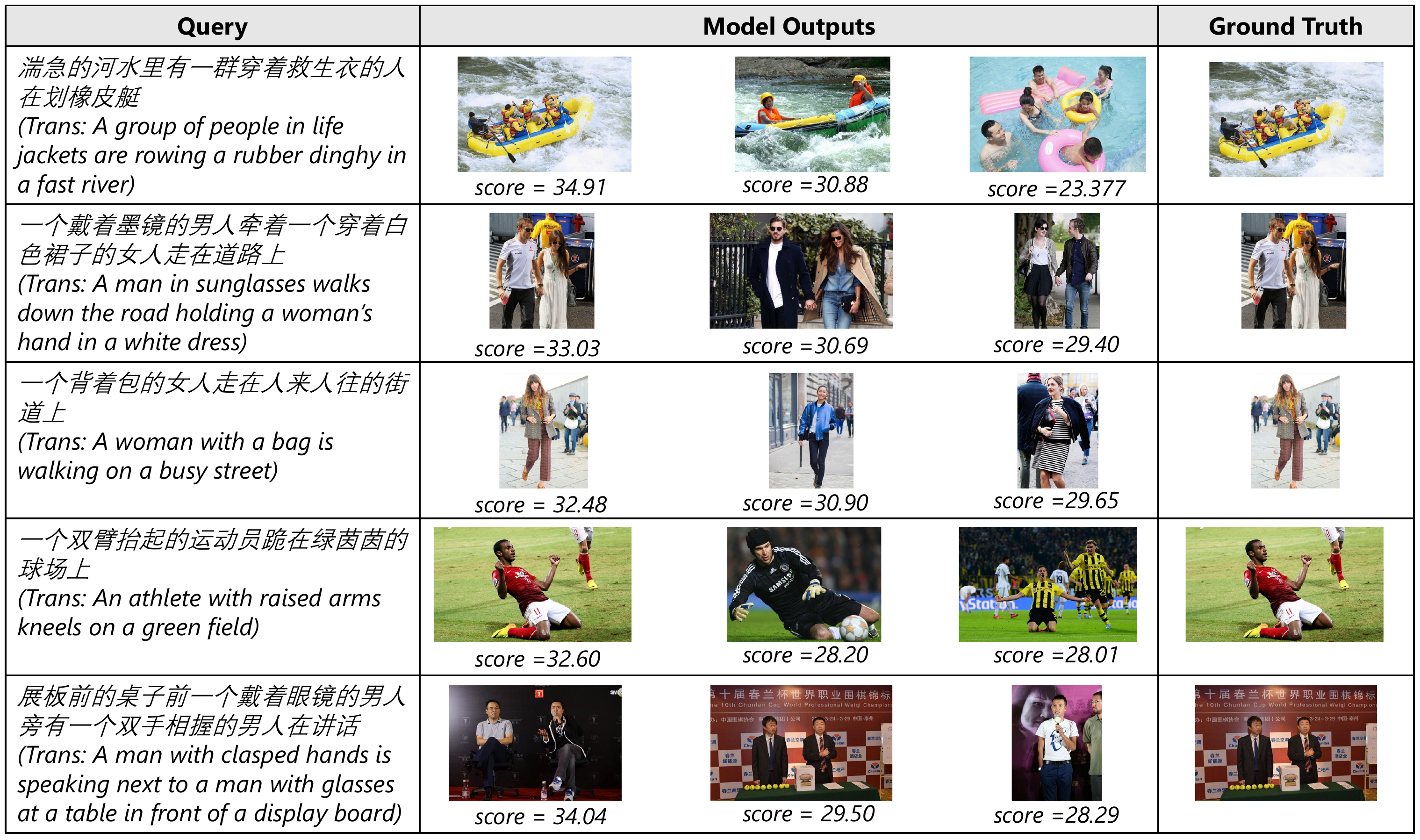}
	\caption{Case study for text-to-image retrieval. For each query, we show top-3 returned images and the relevance scores returned by SkillNet.}
	\label{fig:case-image}
\end{figure}

\begin{figure}[h]
	\centering
	\includegraphics[width=\textwidth]{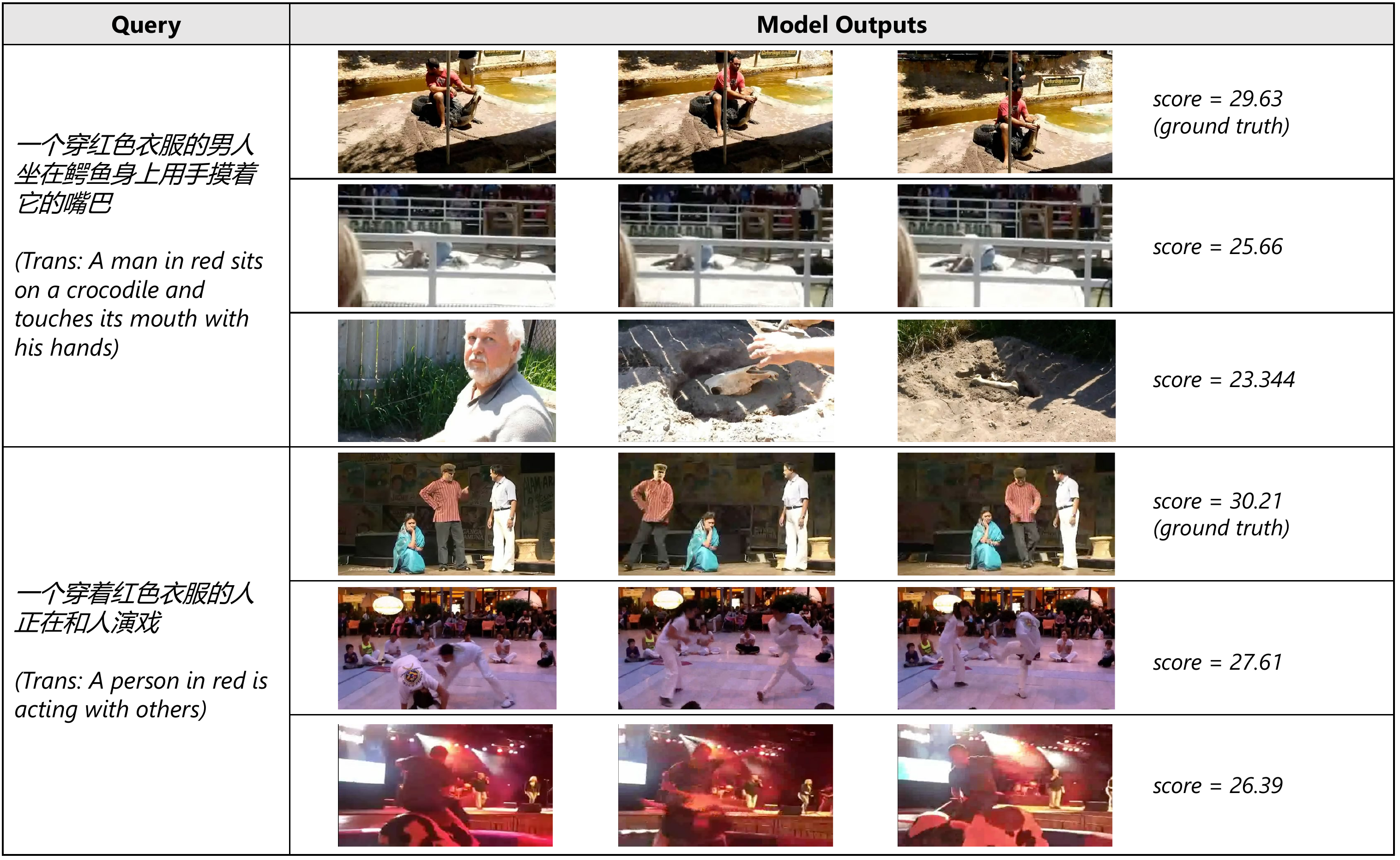}
	\caption{Case study for text-to-video retrieval. For each query, we show the top-3 returned videos returned by SkillNet and provide three frames for each video.}
	\label{fig:case-video}
\end{figure}

\begin{figure}[h]
	\centering
	\includegraphics[width=\textwidth]{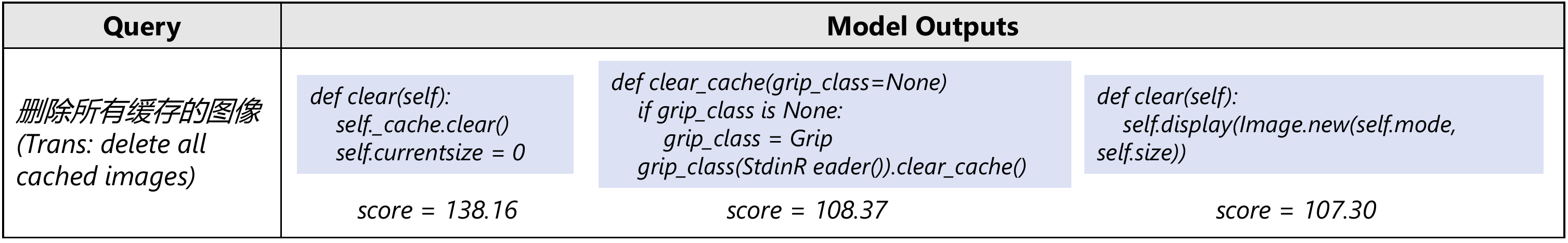}
	\caption{Case study for text-to-code retrieval. For each query, we show top-3 returned codes and the relevance scores returned by SkillNet.}
	\label{fig:case-code}
\end{figure}

Figure \ref{fig:case-image}, \ref{fig:case-video} and \ref{fig:case-code} give case studies on image, video and code retrieval, respectively.

\newpage


\section{Conclusion}
This paper presents a sparsely activated multimodal multitask approach called SkillNet.
We demonstrate the feasibility of using one model to achieve comparable performance compared to multiple modality-specific models.
We further show that sparse pretraining gives a better initialized parameters which leads to improved accuracy, even better than modality-specific pretraining on three of five modalities.
On Chinese text-to-image retrieval, our final system yields better accuracy with less activated parameters compared to existing leading systems.
{Our approach is modality-agnostic and task-agnostic. We leave the extension to larger number of modalities and tasks to the future.}

\small
\setcitestyle{numbers,square}
\bibliographystyle{plainnat}
\bibliography{bibtex}

\appendix
\section{Appendix}
\paragraph{Configurations for Downstream Tasks}

We describe the datasets and configurations for the downstream tasks as described in \S \ref{sec:downstream-tasks}. 
For {\textbf{text}}, we use TNEWS \cite{xu2020clue}, a benchmark dataset for Chinese text classification with 15 categories. It includes 53,300 sentences for training, 10,000 for development, and 10,000 for testing. Evaluation metric is accuracy.
For \textbf{sound}, we adopt the dataset of AISHELL~\cite{bu2017aishell} for automatic speech recognition. It includes 170 hours of speech data in Mandarin. The evaluation metric is character error rate (CER), which means the percentage of characters that are incorrectly predicted (the lower the better).
For \textbf{image}, we use AIC-ICC dataset \cite{wu2017ai}, a benchmark dataset for text-to-image retrieval. It includes 210,000 image-text pairs for training and 30,000 for evaluation. We follow Wukong \cite{gu2022wukong} and consider the first 10,000 images and 50,000 texts from the validation set as the final testing set.
Evaluation metric is Recall @K (e.g., K = 1 and 10).
For \textbf{video}, we carry out text-to-video retrieval on VATEX~\cite{wang2019vatex}, which includes 25,991 videos for training and 3,000 for validation. Since some videos are unavailable for they are deleted or hidden by either YouTube or the users, we actually obtain 23,453 videos for training and 2,709 videos for validation. We randomly select 1,500 videos from validation set as our testing set and use the remaining videos as the development set. In the original dataset, there are 10 sentences in Chinese and 10 sentences in English to describe each video. In this work, we only utilize the Chinese captions. Similar to image retrieval, we use Recall @K (e.g., K = 1, 5 and 10) as the evaluation metrics.
For \textbf{code}, since there is no publicly available dataset for Chinese language, we create a dataset by translating the PyTorrent~\cite{bahrami2021pytorrent} dataset. It contains 218,814 Python package libraries from PyPI and Anaconda environment. We translate English docstrings to Chinese by a translation toolkit Transmart~\cite{huang2021transmart}. We delete duplicate code-text pairs and remove instances with low translation quality. We mix the original training set, development set and test set. At last, we shuffle the mixed set and randomly select 100,000/20,000/30,000 for training, validation and testing, respectively. The evaluation metric is Recall @K.

The model configurations are given as follows. For \textbf{text}, the max length is 512, and a special text padding token is padded if the input is shorter. For \textbf{sound}, we truncate each waveform to no more than 20ms, which leads to the max length of the sound embedding being 1,000. 
If the input is shorter, the remaining positions are filled with a special sound padding token.
Other configurations can be found in \ref{sec:embedding}.  For \textbf{image}, same with ViT-B/16 from CLIP \cite{radford2021learning}, we first resize and normalize each image to $224 \times 224$. Then, we split each image into 196 patches with the patch size of $16 \times 16$, which are sent into a 3 in-channel and 768 out-channel 2D-convolution with kernel size of (16, 16) and stride step of (16, 16). For \textbf{video}, we truncate each video to no more than 10 seconds and transform each video into frames by 3 frames per second. Then, we randomly sample 6 frames for each video. At last, 6 video frames after cropping and normalizing are sent into a 3 in-channel and 768 output-channel 3D-convolution with a kernel size of (3, 16, 16) and stride step (3, 16, 16). For \textbf{code}, we set the whole max length as 512 with the limit of text max length of 64. We use AdamW \cite{loshchilov2018decoupled} optimizer and linear scheduler with 1,000 warmup steps. 
There are different ways to initialize the model parameters. To accelerate the training process, instead of training from random initialization, we use ViT-B/16 from CLIP \cite{radford2021learning} to initialize image-related parameters and initialize other parameters from scratch. 
Since different modalities have different memory costs, we set the batch sizes as 512/1024/3072/1024/512  for text/sound/image/video/code to maximize the memory usage of GPUs.
We observe that sound and code modalities require longer training steps to converge and the data scale of video is smaller than other modalities. Therefore, we sample instances for text/sound/image/video/code modalities with the ratio of 2/4/2/1/4.
For each update, we only sample instances from one modality, which makes the learning process more stable. We update our model for 200,000 steps in total.

\paragraph{Configurations for Pretraining Tasks} We describe the datasets and configurations for the pretraining tasks as described in \S \ref{sec:pre-training}. 
For \textbf{text}, we crawl a collection of raw Chinese texts containing Wikipedia, novels, news, lyrics and poems. We clean the data and finally obtain a dataset of about 300 gigabytes.  
For \textbf{sound}, we collect audio datasets from an open-source platform~\footnote{https://blog.ailemon.net/2018/11/21/free-open-source-chinese-speech-datasets/}, which includes about 1,200 hours of Chinese speech data. 
For \textbf{image}, we download the Wukong dataset \cite{gu2022wukong} which originally includes 101,483,885 text-image pairs and filter out low-quality instances that with no Chinese words, too many illegal symbols and the length of captions is less than 4. We finally use about 84,000,000 text-image pairs for pretraining. 
For \textbf{video}, we use WebVid-2M~\cite{bain2021frozen}, which comprises of over two million video-text pairs scraped from the internet. We translate the original English texts to Chinese by the translation tool Transmart and use the translated data for pretraining.
For \textbf{code} pretraining, we hold out 800,000 code-text pairs from the aforementioned code dataset translated from PyTorrent, which have no overlaps with the datasets used for the downstream task of text-code retrieval. 

We pretrain SkillNet using the AdamW \cite{loshchilov2018decoupled} optimizer with the learning rate 1e-5 and a linear scheduler with 10,000 warmup steps. Same with the configuration for downstream tasks, we use one-modality data to pretrain our model in each update. The model is pretrained for 1,000,000 steps in total with batch sizes of 1024/512/8192/2048/512 for text/sound/image/video/code, respectively. 
Pretraining takes about 14 days on 64 A100 GPUs of 40GB memory size.

\end{document}